# Point Cloud Context Analysis for Rehabilitation Grasping Assistance

Jackson M. Steinkamp
Harvard College
Cambridge, MA

Laura J. Brattain
MIT Lincoln Laboratory
Lexington, USA

Conor J. Walsh
and Robert D. Howe
School of Engineering and
Applied Sciences
Harvard University
Cambridge, MA

*Abstract*—Controlling hand exoskeletons for assisting impaired patients in grasping tasks is challenging because it is difficult to infer user intent. We hypothesize that majority of daily grasping tasks fall into a small set of categories or modes which can be inferred through real-time analysis of environmental geometry from 3D point clouds. This paper presents a low-cost, real-time system for semantic image labeling of household scenes with the objective to inform and assist activities of daily living. The system consists of a miniature depth camera, an inertial measurement unit and a microprocessor. It is able to achieve 85% or higher accuracy at classification of predefined modes while processing complex 3D scenes at over 30 frames per second. Within each mode it can detect and localize graspable objects. Grasping points can be correctly estimated on average within 1 cm for simple object geometries. The system has potential applications in robotic-assisted rehabilitation as well as manual task assistance.

## I. Introduction

Rehabilitation robotics offers the possibility of augmenting standard therapy programs by supporting, assisting, and structuring movements in task-specific ways, such as passive assistance by relieving the weight of the arm during reaching tasks [1, 2] or by actively providing assistance of hand and finger function [3, 4, 5, 6]. Recently there has been a growing interest in using exoskeletons to help with activities of daily living (ADLs) [7, 8, 9, 10]. These devices use wearable actuators to apply force to fingers to help with grasping. A main issue is that impaired users have trouble controlling fingers, so conventional control modes, such as using EMG sensing of muscle activation of the muscles in the forearm [11], may be inadequate to control actuators to produce correct grasping behavior. Similarly, controlling grasping behavior in autonomous robots in unstructured environments is challenging because of the complexity of the cluttered setting and the variability of target objects. [12, 13].

A promising approach to augmenting exoskeleton control is the use of vision-based methods. In particular, depth cameras have become ubiquitous in robotic sensing and have many advantages over RGB vision for acquiring 3D geometry. There has been an influx of work related to indoor environment scene segmentation, scene modeling, semantic labeling and support inference from RGB-D cameras [14, 15, 16, 17, 18], or from 3D point cloud maps alone [19, 20, 13, 21].

We propose the use of wrist-mounted video-plus-depth cameras and inertial measurement units (IMUs) to characterize the surrounding environment, and thus determine both key object properties and the anticipated task category. In this scenario, the arm moves the hand towards the target object, while the camera acquires point clouds of the approaching object. If processing algorithms determine that the surrounding environment is largely a horizontal plane (e.g. table top) for example, this indicates that the likely goal is grasping and lifting the object. If the object is protruding from vertical surface, it is likely a door knob or handle, and should be grasped to permit turning and pulling or pushing. Similar combinations of environmental configurations can be used to infer control modes for keypad button pushing, crowded shelves, and other common contexts seen in ADLs.

The goal of this paper is to explore scene context analysis, including control mode detection, for the purpose of inferring user intent and assisting with grasping. We demonstrate our concept with an initial system prototype. We begin by describing the system design, followed by algorithms on scene parsing and mode detection. We then present our preliminary results and conclude with discussion and future work.

## II. Materials and Methods

### A. Overall System Design

Our system, as shown in Figure 1, consists of a miniature 3D depth camera for sensing environmental geometry, an IMU for detecting hand and camera orientation, and a microprocessor to enable communication between the IMU and the computer. The system is designed to sense and classify contextual properties of a scene in order to inform desired hand function. We developed algorithms to accurately and reliably classify daily household scenes at real-time frame rate ($\geq$30 fps), and then use the semantic labels to guide further scene parsing in order to detect objects and identify potential grasp heuristics. The capabilities can be used in task assistance through integrating with a wearable rehabilitation system or with other robotic systems to provide scene context classifications and grasping point estimations.

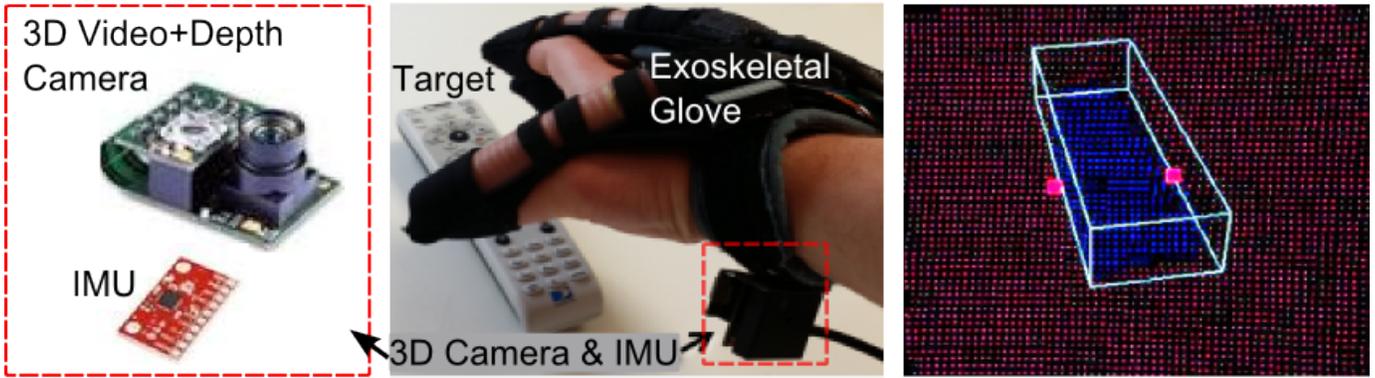

Fig. 1. Real-time 3D object sensing. (*Left*) 3D depth+video camera and IMU; (*Middle*) Exoskeletal glove mounted with a depth camera and IMU looking at the target object; (*Right*) Processed depth image: red pixels = table top, blue = object; white = bounding box of object from segmented object and PCA; pink dots = target grasp points determined as opposition points on minor axis parallel to table.

## B. Sensors

To sense the environment, the system uses a wrist-mounted depth camera to capture 3D scenes. The CamBoard Nano (pmdtechnologies.com, Siegen, Germany) proved ideal for this purpose, as it is a small, low-cost depth camera which outputs accurate depth readings between 5 and 100 cm. The Nano provides better resolution, closer range, faster frame rate, and smaller footprint than the Microsoft Kinect (Microsoft Corporation, Redmond, WA, USA), the de facto standard depth camera for 3D scene analysis. The vast majority of relevant environmental features fall within the 5 to 100 cm range as the hand reaches for an object.

In order to distinguish horizontal planes from vertical planes and to conduct basic odometry of the camera's motion, we attached a 6-axis IMU (MPU-6050, InvenSense, San Jose, CA, USA) to the back of the camera. It consists of a three-axis gyroscope and a three-axis accelerometer, and is widely used for orientation sensing in robotic applications. Figure 2 describes the key system modules and the data flow and communication.

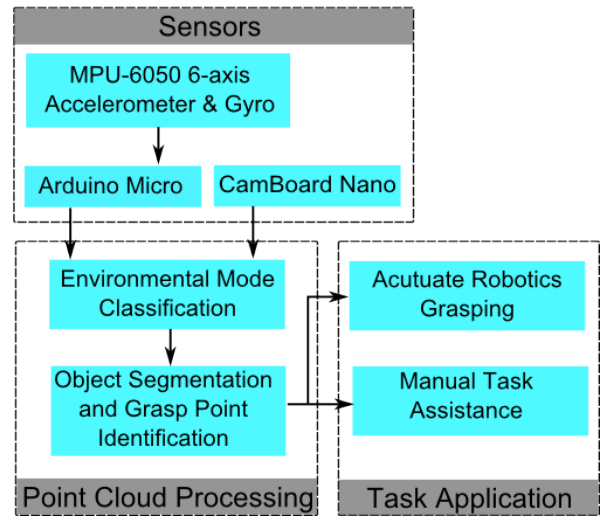

Fig. 2. System modules and data flow.

## C. Mode Detection and Algorithms

Selecting a proper set of detectable grasp contexts is a critical design choice. In this initial prototype, we have focused on a few modes which are highly important to daily life activities. Modes which we have so far implemented include a "tabletop" object mode, which detects objects resting on horizontal planes, a "doorknob" mode, which detects unsupported objects extruding from vertical planes, and "pot handle" mode, which detects elongated objects not directly supported by planes. Despite their specific titles, this small suite of modes is sufficient and applicable to a surprisingly large variety of household grasping tasks.

Since we are working with three-dimensional scenes, computational efficiency is also a concern. We chose to use the open-source Point Cloud Library (PCL) [22, 23] and the point cloud format for representing 3D images. The point cloud format, which represents a 3D image as an array of points rather than as a two-dimensional array of z-coordinates, is the natural choice for three-dimensional operations such as nearest neighbor detection, surface normal estimation, and three-dimensional surface and shape detection. In addition, the PCL provides a standard, optimized suite of state-of-the-art functions and algorithms for point cloud manipulation.

The algorithms developed here are geared towards real-time analysis of environment geometry in order to identify the context of the grasping task. As shown in Figure 3, depth images are captured from the CamBoard Nano at a framerate of 60 Hz and converted from range images into point cloud format. Initial preprocessing steps remove pixels flagged by the camera as inaccurate, saturated, or inconsistent, and convert each valid pixel to a 3D point, and then store all of the points in a single point cloud. After this, we perform context extraction. By using random sample consensus (RANSAC), we detect the most prominent plane in the point cloud, as measured by the number of points in the cloud. RANSAC proved robust at detecting planes even when the depth image was cluttered or the planes were obscured by other objects. It proved to be more effective than Hough transforms for identifying planes with a large number of possible orientations [24]. Next, accelerometer data is used to determine the orientation of the plane; the camera and accelerometer share a fixed known rigid transform, allowing estimation of the camera's orientation with respect to world coordinates. Two special cases are horizontal and vertical planes. Here we assume that a highly prominent horizontal plane implies a tabletop scene and a highly prominent vertical plane infers a doorknob scene, or possibly a remote control keypad or microwave button keypad scene. If the ratio between the number of RANSAC plane outliers and inliers in a scene is below a predefined threshold, then there exists a prominent plane. This also serve as a confidence metric when comparing scenes with multiple possible modes.

If no prominent plane exists in the scene, we search for another possible mode - pot handle mode. The criteria for this involves a prominent cluster of points with a small minor axis, above a plane but not directly supported by it.

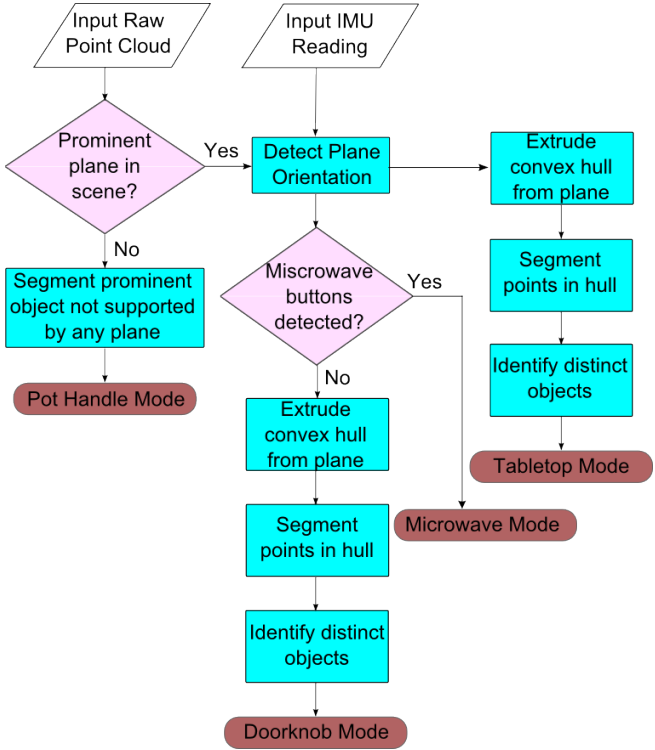

Fig. 3. Algorithm flowchart describing 3D scene parsing and mode detection

### D. Scene Analysis

Once the mode is identified, we can then further parse the relevant scene based on the mode to extract key information. This both minimizes superfluous computation and targets grasp heuristic analysis and object segmentation techniques. For instance, it is useful to estimate desirable grasp locations on a target object, i.e. points where fingers can be placed to grasp and lift the object. These locations can be used by an exoskeleton assist glove controller to determine when to help close the user's fingers.

Our prototype system used a simple heuristic that desirable grasp points are often located at the center of the sides of an object sitting on a table. If the system identifies a prominent horizontal plane in the scene, we proceed by identifying the convex hull of the plane and extruding it along the surface normal in order to identify the volume which could potentially contain graspable tabletop objects. Then, we segment point clusters inside the convex hull in order to identify distinct objects. From there, we use principal components analysis (PCA) to identify the principal axes of the object, estimate a bounding box for the object, and find a minor axis and opposing grasping points along that axis. The grasping points are calculated by finding the nearest point to the minor axis on the object's surfaces perpendicular to the planar surface the object extrudes from. A similar procedure is used to identify objects extruding from vertical surfaces, such as doorknobs.

In the case of pot handle mode, we look for point clusters greater than a certain height (e.g. 5 cm) above a plane. In the case of the keypad mode detection, if the system identifies an array of keypad buttons as the salient environmental feature, we will use standard 2D computer vision techniques such as a Haar Cascade or HOG features in combination with a Support Vector Machine, for example, to locate buttons.

The key insight here is that identification of the proper environmental context determines the subsequent steps for object identification and grasping point estimation. As a result detailed analysis is only focused on the relevant parts of the scene. This greatly reduces computation time.

## III. EXPERIMENTS AND RESULTS

The goal of this study is to demonstrate the feasibility of the approach and explore implementation methods. To this end, we conducted experiments to measure the performance in terms of both mode classification and grasp point identification.

### A. Mode Classification

To test mode classification, we performed bench top experiments in various scenes, including a crowded kitchen scene and a computer desk scene. The scenes included doorknobs, planar surfaces with multiple orientations, and numerous tabletop objects. Performance was measured by comparing the mode output by the system to the correct mode as identified in the previous sections. Video was recorded from various views, positions, camera orientations, and with different object placements.

Figure 4 is an example of a horizontal plane or tabletop scene. An RGB image of the tabletop scene is shown on the top, and the processed point cloud image of the same scene is shown at the bottom. The dark red points represent the detected prominent horizontal plane in the scene, and the blue points represent the identified tabletop objects. The bounding box of each object is outlined in white lines, and the estimated grasping points are identified as the light red points on the bounding box.

Figure 5 is an example of a vertical plane scene of a door and its handle. To the left is the RGB image of the scene, and to the right is the processed point cloud of the same scene. The dark red points are those identified by the algorithm as part of the vertical plane, and the bounding box is drawn in white lines with estimated grasping points shown in light red.

Similarly, Figure 6 shows a pot handle scene which includes an RGB image of the pot and the corresponding processed point cloud image with the estimated pot handle bounding box and grasping point.

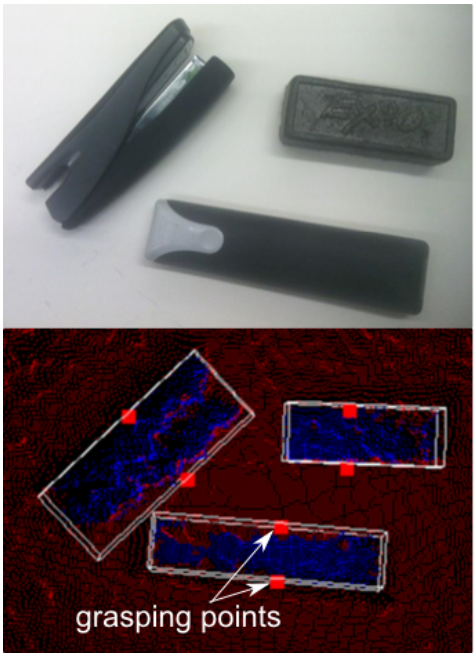

Fig. 4. Example tabletop scene. (*Top*) RGB image of a tabletop scene; (*Bottom*) Processed 3D point cloud with grasping points identified. The bounding box for each of the detected object is displayed in white outlines.

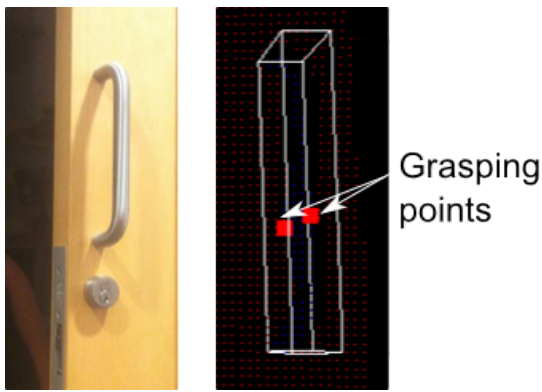

Fig. 5. Example doorknob scene. (*Left*) RGB image of a doorknob scene; (*Right*) Processed 3D point cloud with object bounding box and grasping point identified.

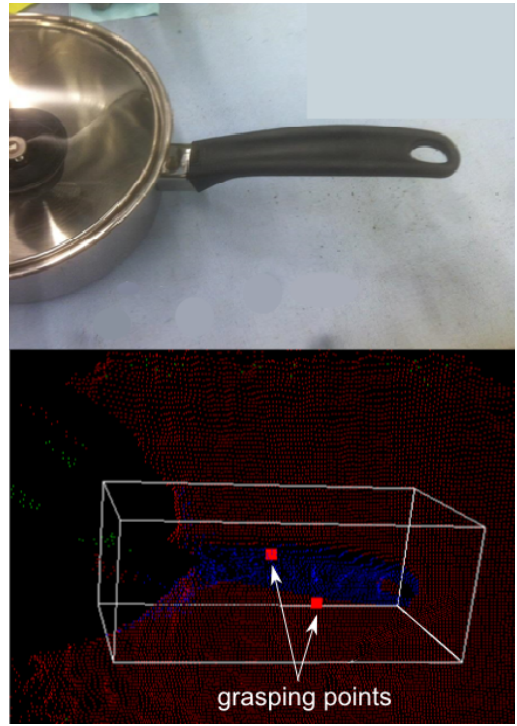

Fig. 6. Example pot handle scene. (*Top*) RGB image of a pot handle scene; (*Bottom*) Processed 3D point cloud with object bounding box and grasping points identified.

We conducted a number of five minute real-time experiments in different household scenes as described above. Our system was able to correctly detect prominent planes, even when as much as 80% of the visible table surface was occluded by tabletop objects. In terms of scene classification, in the crowded kitchen scene and computer desk scene, tabletop and doorknob mode were detected correctly 100% of the time. Pot handle mode was detected correctly 85% of the time.

### B. Grasping Point Accuracy

To evaluate the performance of the model in estimating grasping points, we tested it on numerous household objects encountered in ordinary activities of daily living. Although a simple grasping heuristic was used (opposing points centered along the minor axis), the system proved to perform well at estimating grasping points on a variety of objects as shown in the scenes above.

To give a measure of accuracy of the grasping point detection method, we compared the distance of each pair of estimated grasping points to the actual distance measurement of the two opposing centers of a rectangular object (a white board eraser) along its minor axis using a caliper (0.25 mm accuracy). We calculated the accuracy at different hand-object distances in order to simulate a hand approaching an object for grasping. We looked at the operable range of the depth camera in which the whole object was visible.

Figure 7 is a plot of the grasping point detection accuracy. As the camera approaches the object from the top, the system is able to correctly localize opposing grasping points to within 5-6 mm of their actual positions. Error decreases approximately linearly with distance, and the majority of the error in grasping point detection comes from the z-axis error, due to the depth camera's top-down perspective as it approaches the object. The grasp point location estimates depend on sensing both the upper surface of the object and the table top height; the actual vertical width of the object is sometimes occluded by the points on the upper surface of the eraser.

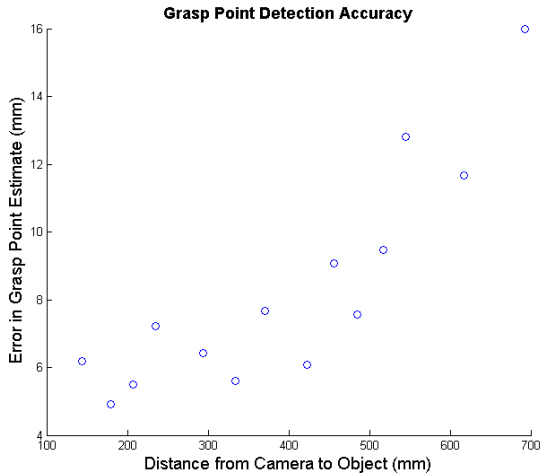

Fig. 7. Accuracy of estimated grasping points at different distances where the object was visible.

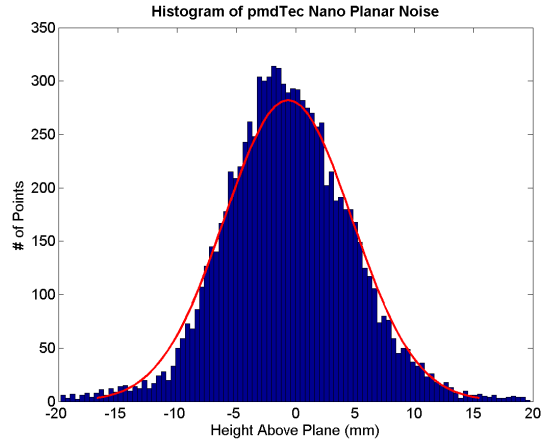

Fig. 8. Histogram of planar noise in the CamBoard Nano depth camera overlayed with best-fit Gaussian PDF

*C. Noise Measurements*

We also evaluated the performance and limitations of our chosen hardware, notably, the depth camera. The CamBoard Nano, while providing effective depth images, is not specified to work well on reflective surfaces due to low intensity of the returned infrared signal. Highly reflective objects therefore post a challenge to our system. Similarly, even when looking at a perfectly planar surface, the depth camera shows wavelike noise patterns with standard deviation of 5-8 mm (Figure 8), as expected for time-of-flight depth sensing cameras. This prevents the object detection algorithms from correctly distinguishing small ($\leq 1$ cm) objects resting on planar surfaces. Fortunately, because the noise can be well-modeled as Gaussian, simple statistical signal processing techniques combined with spatial coherence estimates can be used to detect small objects.

The noise also varies based on the depth camera's distance from a plane, as shown in Figure 9. The noise appears to be approximately Gaussian in the 30-80 cm range, enabling use of the above- mentioned signal processing methods.

Other measurements further establish the performance and limitations of our system. Through experiments, we found that our system can work robustly even with 80% of a planar surface occluded by clutter. In addition, we found that two planes, one supporting the other (e.g. a stack of books), are distinguished from each other when the upper plane has a height greater than 1.9 cm above the lower plane. These measurements will help fine-tune the RANSAC threshold parameters and further improve system performance.

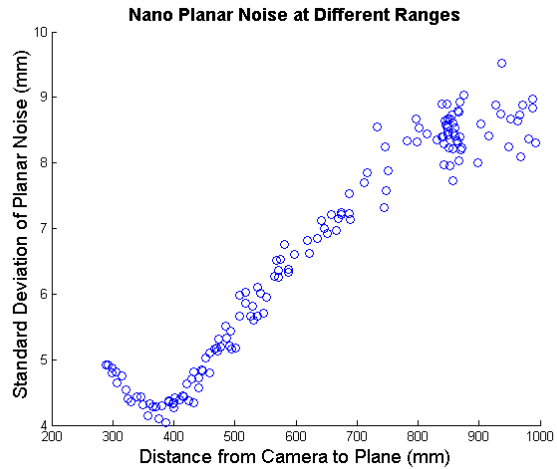

Fig. 9. Standard deviation of planar noise at different ranges showing that the CamBoard Nano has predictable noise behavior between 30-80 cm.

IV. DISCUSSION

In this paper, we presented an initial prototype system for grasping task context detection, using a miniature depth camera and an IMU mounted at the user's wrist. The results show that such a system can provide low-cost, real-time classification of scenes into informative categories, which can be used to aid users in controlling a grasping-assistance exoskeleton. For example, when the user moves their hand towards an object resting on a flat horizontal surface, the system can infer that the goal is to grasp the object, and trigger the fingers to close when the hand is poised above the object, with the target grasp locations near the fingers. Many tasks of daily living, for example, picking up a coffee cup or pushing a button on a microwave keypad fall within the three modes implemented here. Overall, such a system has great potential in task assistance and rehabilitation.

*A. Extensions*

The system is easily extensible to include a larger library of semantic labels, because environmental geometry is often effective in identifying these categories. In addition, the system naturally produces large, high-resolution close-up point clusters because the user moves their hand towards the objects they wish to interact with. By focusing computation and analysis on these features, we minimize the computation cost. For example, as a hand moves toward an object continually, as detected by the combination of the IMU and the depth camera, we can focus scene processing on that particular object, because the user likely intends to grasp it - or at least, the user cannot grasp an object that the hand is not approaching. As a result, not only does our scene processing inform the grasping task, the grasping task also informs our scene processing. Scene prominence and object size are useful proxies for user intent.

This system is particularly appropriate for stroke survivors with hand impairments. These individuals often have good residual control of shoulder and elbow joints, but poor control of the fingers. Traditional methods of exoskeleton control like surface EMG signals are problematic for the hand, due to the large number of overlapping muscles in the forearm. Because it uses environmental cues, our depth camera based system can minimize the need for detailed control inputs from the user. However, effective use in real-world rehabilitation and assistance applications will doubtless require tight integration of the proposed system with a number of sensing and control methods.

*B. Future Work*

The depth camera's noise is a factor in the performance of our system, as the planar noise level is directly related to the RANSAC planar segmentation threshold that we set. For instance, if the depth camera is noisier, the plane detection algorithm must be more lenient (larger threshold value), making it more difficult to distinguish planar noise from objects with small height resting on the planar surface. In our next phase of development, we will apply appropriate filtering techniques based on the statistical measures of system performance to improve planar identification and object differentiation.

One untapped performance enhancement is the use of frame-to-frame registration and sensor histories. Capturing multiple views of the scene would both increase the robustness of the segmentation algorithms as well as provide increased grasping point detection accuracy due to the more accurate object models created by registration. Existing registration libraries could be leveraged to add frame-to-frame scene construction. Similarly, existing libraries for simultaneous localization and mapping (SLAM) would allow for better object model construction and thus better decision making [25]. However, real-time SLAM is challenging in the context of grasping, as scenes change at a rate much more quickly than in the navigation contexts which are the basis for the bulk of existing SLAM research. In the future, we will investigate the integration of SLAM with the depth camera for more robust scene registration and model construction, while still preserving real-time performance, which is critical for applications in task assistance.

Further work is required to determine the limits of performance of the approach. One key question will be the degree to which the detected contexts always correspond to specific task intents. Users will be frustrated if, for example, the exoskeleton system incorrectly attempts to grasp a remote control when the intent was to push a button or vice versa, but even a small level of user input may be sufficient to correct these issues. It is also unclear how well the system can identify the appropriate grasp configuration for a given object; while object size is very often a good heuristic (pinch grasps for small objects, wrap grasp for large, etc.), grasp choice also depends on the intended task as well as the object. The next step in this project is to connect the depth camera system to a glove exoskeleton [9] to enable exploration of the behavior of the system in real-world settings, so such questions can begin to be addressed.

We will also work to expand the suite of environmental contexts to give the system the ability to handle a large variety of daily household tasks. Example of future modes to be developed include a clothing mode, which detects highly irregular folded surfaces, either hanging vertically or resting horizontally, and a shelf mode - an unsupported horizontal edge, indicating a horizontal surface, but with too much clutter on it to detect a plane.

ACKNOWLEDGMENT

The authors would like to thank Dr. Panagiotis Polygerinos at Wyss Institute at Harvard University for his assistance on the exoskeletal glove. Laura Brattain's work is sponsored by the Department of the Air Force under Air Force contract FA8721-05-C-0002. Opinions, interpretations, conclusions and recommendations are those of the authors and are not necessarily endorsed by the United States Government.

REFERENCES

[1] B. T. Volpe, H. I. Krebs, N. Hogan, L. E. OTR, C. Diels, and M. Aisen, "A novel approach to stroke rehabilitation: robot-aided sensorimotor stimulation," *Neurology*, vol. 54, no. 10, pp. 1938–1944, May 23 2000, lR: 20071115; GR: HD 37397/HD/NICHD NIH HHS/United States; JID: 0401060; ppublish.

[2] S. J. Housman, K. M. Scott, and D. J. Reinkensmeyer, "A randomized controlled trial of gravity-supported, computer-enhanced arm exercise for individuals with severe hemiparesis," *Neurorehabilitation and neural repair*, vol. 23, no. 5, pp. 505–514, Jun 2009, jID: 100892086; 2009/02/23 [aheadofprint]; ppublish.

[3] C. D. Takahashi, L. Der-Yeghiaian, V. Le, R. R. Motiwala, and S. C. Cramer, "Robot-based hand motor therapy after stroke," *Brain : a journal of neurology*, vol. 131,

no. Pt 2, pp. 425–437, Feb 2008, gR: 5M011 RR-00827-29/RR/NCRR NIH HHS/United States; JID: 0372537; 2007/12/20 [aheadofprint]; ppublish.

[4] H. C. Fischer, K. Stubblefield, T. Kline, X. Luo, R. V. Kenyon, and D. G. Kamper, "Hand rehabilitation following stroke: a pilot study of assisted finger extension training in a virtual environment," *Topics in Stroke Rehabilitation*, vol. 14, no. 1, pp. 1–12, 2007.

[5] L. Dovat, O. Lambercy, R. Gassert, T. Maeder, T. Milner, T. C. Leong, and E. Burdet, "Handcare: a cable-actuated rehabilitation system to train hand function after stroke," *Neural Systems and Rehabilitation Engineering, IEEE Transactions on*, vol. 16, no. 6, pp. 582–591, 2008.

[6] X. Luo, T. Kline, H. C. Fischer, K. A. Stubblefield, R. V. Kenyon, and D. G. Kamper, "Integration of augmented reality and assistive devices for post-stroke hand opening rehabilitation," in *Engineering in Medicine and Biology Society, 2005. IEEE-EMBS 2005. 27th Annual International Conference of the*. IEEE, 2005, pp. 6855–6858.

[7] L. Connelly, M. E. Stoykov, Y. Jia, M. L. Toro, R. V. Kenyon, and D. Kamper, "Use of a pneumatic glove for hand rehabilitation following stroke," in *Engineering in Medicine and Biology Society, 2009. EMBC 2009. Annual International Conference of the IEEE*. IEEE, 2009, pp. 2434–2437.

[8] P. M. Aubin, H. Sallum, C. Walsh, L. Stirling, and A. Correia, "A pediatric robotic thumb exoskeleton for at-home rehabilitation: The isolated orthosis for thumb actuation (iota)," in *Rehabilitation Robotics (ICORR), 2013 IEEE International Conference on*. IEEE, 2013, pp. 1–6.

[9] P. Polygerinos, S. Lyne, Z. Wang, L. F. Nicolini, B. Mosadegh, G. M. Whitesides, and C. J. Walsh, "Towards a soft pneumatic glove for hand rehabilitation," in *Intelligent Robots and Systems (IROS), 2013 IEEE/RSJ International Conference on*. IEEE, 2013, pp. 1512–1517.

[10] B. Mosadegh, P. Polygerinos, C. Keplinger, S. Wennstedt, R. F. Shepherd, U. Gupta, J. Shim, K. Bertoldi, C. J. Walsh, and G. M. Whitesides, "Pneumatic networks for soft robotics that actuate rapidly," *Advanced Functional Materials*, vol. 24, no. 15, pp. 2163–2170, 2014.

[11] L. Lucas, M. DiCicco, and Y. Matsuoka, "An emg-controlled hand exoskeleton for natural pinching," *Journal of Robotics and Mechatronics*, vol. 16, pp. 482–488, 2004.

[12] Z.-C. Marton, L. Goron, R. B. Rusu, and M. Beetz, *Reconstruction and verification of 3D object models for grasping*, ser. Robotics Research. Springer, 2011, pp. 315–328.

[13] R. B. Rusu, Z. C. Marton, N. Blodow, M. Dolha, and M. Beetz, "Towards 3d point cloud based object maps for household environments," *Robotics and Autonomous Systems*, vol. 56, no. 11, pp. 927–941, 2008.

[14] D. Filliat, E. Battesti, S. Bazeille, G. Duceux, A. Gepperth, L. Harrath, I. Jebari, R. Pereira, A. Tapus, and C. Meyer, "Rgbd object recognition and visual texture classification for indoor semantic mapping," in *Technologies for Practical Robot Applications (TePRA), 2012 IEEE International Conference on*. IEEE, 2012, pp. 127–132.

[15] P. Henry, M. Krainin, E. Herbst, X. Ren, and D. Fox, "Rgb-d mapping: Using depth cameras for dense 3d modeling of indoor environments," in *In the 12th International Symposium on Experimental Robotics (ISER*. Citeseer, 2010.

[16] D. Holz, S. Holzer, R. B. Rusu, and S. Behnke, *Real-time plane segmentation using RGB-D cameras*, ser. RoboCup 2011: Robot Soccer World Cup XV. Springer, 2012, pp. 306–317.

[17] H. S. Koppula, A. Anand, T. Joachims, and A. Saxena, "Semantic labeling of 3d point clouds for indoor scenes," in *Advances in Neural Information Processing Systems*, 2011, pp. 244–252.

[18] N. Silberman, D. Hoiem, P. Kohli, and R. Fergus, *Indoor segmentation and support inference from RGBD images*, ser. Computer Visionâ"ECCV 2012. Springer, 2012, pp. 746–760.

[19] Y. Cui, S. Schuon, D. Chan, S. Thrun, and C. Theobalt, "3d shape scanning with a time-of-flight camera," in *Computer Vision and Pattern Recognition (CVPR), 2010 IEEE Conference on*. IEEE, 2010, pp. 1173–1180.

[20] R. Fabio, "From point cloud to surface: the modeling and visualization problem," *International Archives of Photogrammetry, Remote Sensing and Spatial Information Sciences*, vol. 34, no. 5, p. W10, 2003.

[21] R. B. Rusu, Z. C. Marton, N. Blodow, A. Holzbach, and M. Beetz, "Model-based and learned semantic object labeling in 3d point cloud maps of kitchen environments," in *Intelligent Robots and Systems, 2009. IROS 2009. IEEE/RSJ International Conference on*. IEEE, 2009, pp. 3601–3608.

[22] R. B. Rusu and S. Cousins, "3d is here: Point cloud library (pcl)," in *Robotics and Automation (ICRA), 2011 IEEE International Conference on*. IEEE, 2011, pp. 1–4.

[23] A. ten Pas and R. Platt, "Localizing grasp affordances in 3-d points clouds using taubin quadric fitting," *arXiv preprint arXiv:1311.3192*, 2013.

[24] R. Schnabel, R. Wahl, and R. Klein, "Efficient ransac for pointâcloud shape detection," in *Computer graphics forum*, vol. 26. Wiley Online Library, 2007, pp. 214–226.

[25] Y. Taguchi, Y.-D. Jian, S. Ramalingam, and C. Feng, "Slam using both points and planes for hand-held 3d sensors," in *Mixed and Augmented Reality (ISMAR), 2012 IEEE International Symposium on*. IEEE, 2012, pp. 321–322.